\documentclass{article}
\PassOptionsToPackage{dvipsnames,table}{xcolor}
\usepackage{iclr2027_conference,times}

\usepackage{amsmath,amsfonts,bm}

\def\eqref#1{equation~\ref{#1}}

\def\1{\bm{1}}

\DeclareMathAlphabet{\mathsfit}{\encodingdefault}{\sfdefault}{m}{sl}
\SetMathAlphabet{\mathsfit}{bold}{\encodingdefault}{\sfdefault}{bx}{n}

\usepackage{float}
\usepackage{hyperref}
\usepackage{url}
\usepackage{graphicx}
\usepackage{booktabs}
\usepackage{xspace}
\usepackage{tikz}
\usepackage{multirow}
\usepackage{adjustbox}
\usepackage{pifont}
\usepackage{subcaption}
\usepackage{adjustbox}
\usepackage{enumitem}
\usepackage{wrapfig}
\usepackage{caption}
\usepackage{placeins}
\definecolor{colFull}{HTML}{D6EAF8}   
\definecolor{colLoRA}{HTML}{D9F2D0}   
\definecolor{colLoKr}{HTML}{FFFFD5}   
\definecolor{colLoHa}{HTML}{CAEEFB}   
\definecolor{colDelta}{HTML}{EAF0FB}

\definecolor{ReferenceGray}{HTML}{F7F7F3}
\definecolor{QuantLavender}{HTML}{F2EFF9}
\definecolor{PruneGreen}{HTML}{EEF6F0}
\definecolor{JointBlue}{HTML}{E9F2FA}
\definecolor{RelAccBlue}{HTML}{315C99}

\makeatletter
\renewcommand{\hyper@natlinkbreak}[2]{#1}
\makeatother

\title{4D-GSW: Kinematic-Aware Spatio-Temporal Consistent Watermarking for 4D Gaussian Splatting}

\author{%
Sifan Zhou$^{1}$\thanks{Equal contribution.~(sifanjay@gmail.com)}, \hspace{.3em}
Hang Zhang$^{2}$\footnotemark[1], \hspace{.3em}
Yuhang Wang$^{3}$, \hspace{.3em}
\textbf{Ming Li$^{4}$}\thanks{{}{}Corresponding authors (ming.li@u.nus.edu).} \hspace{.3em}\\
[1ex]
$^{1}$School of Automation, Southeast University \\
$^{2}$Guangdong Laboratory of Artificial Intelligence and Digital Economy (Shenzhen) \\
$^{3}$University of Chinese Academy of Sciences \\
$^{4}$School of Artificial Intelligence, Chinese University of Hong Kong (Shenzhen)\\
}

\iclrfinalcopy
\begin{document}

\maketitle

\begin{abstract}
While 4D Gaussian Splatting (4DGS) has revolutionized high-fidelity dynamic reconstruction, safeguarding the intellectual property of these assets remains an open challenge. Conventional steganographic techniques often neglect the underlying kinematic manifolds, triggering non-physical artifacts such as severe temporal flickering and "FVD collapse". To address this, we propose \textbf{4D-GSW}, a kinematic-aware watermarking framework designed to embed robust copyright information while preserving high spatio-temporal consistency. Unlike prior 4D steganography that primarily focuses on opacity-guided invisibility, our approach explicitly addresses the physical coherence of motion trajectories. We introduce a \textbf{Spatio-Temporal Curvature (STC)} metric to identify "Dynamic Instants," adaptively gating watermark gradient injection to shield critical motion manifolds from non-physical perturbations. To ensure global coherence across complex deformations, we formulate a joint \textbf{HMM-MRF energy minimization} model that synchronizes watermark phases within both temporal trajectories and spatial neighborhoods. Furthermore, an \textbf{anisotropic gradient routing} mechanism ensures that watermark embedding reduces interference with photometric reconstruction. Extensive experiments have demonstrated the superior performance of our method in robustly hiding watermarks while resisting various attacks and maintaining high rendering quality and spatiotemporal consistency.
\end{abstract}
\vspace{-4mm}
\section{Introduction}
\label{Introduction}
\vspace{-2mm}
\begin{figure}[!t]
\centering
\includegraphics[width=0.75\linewidth]{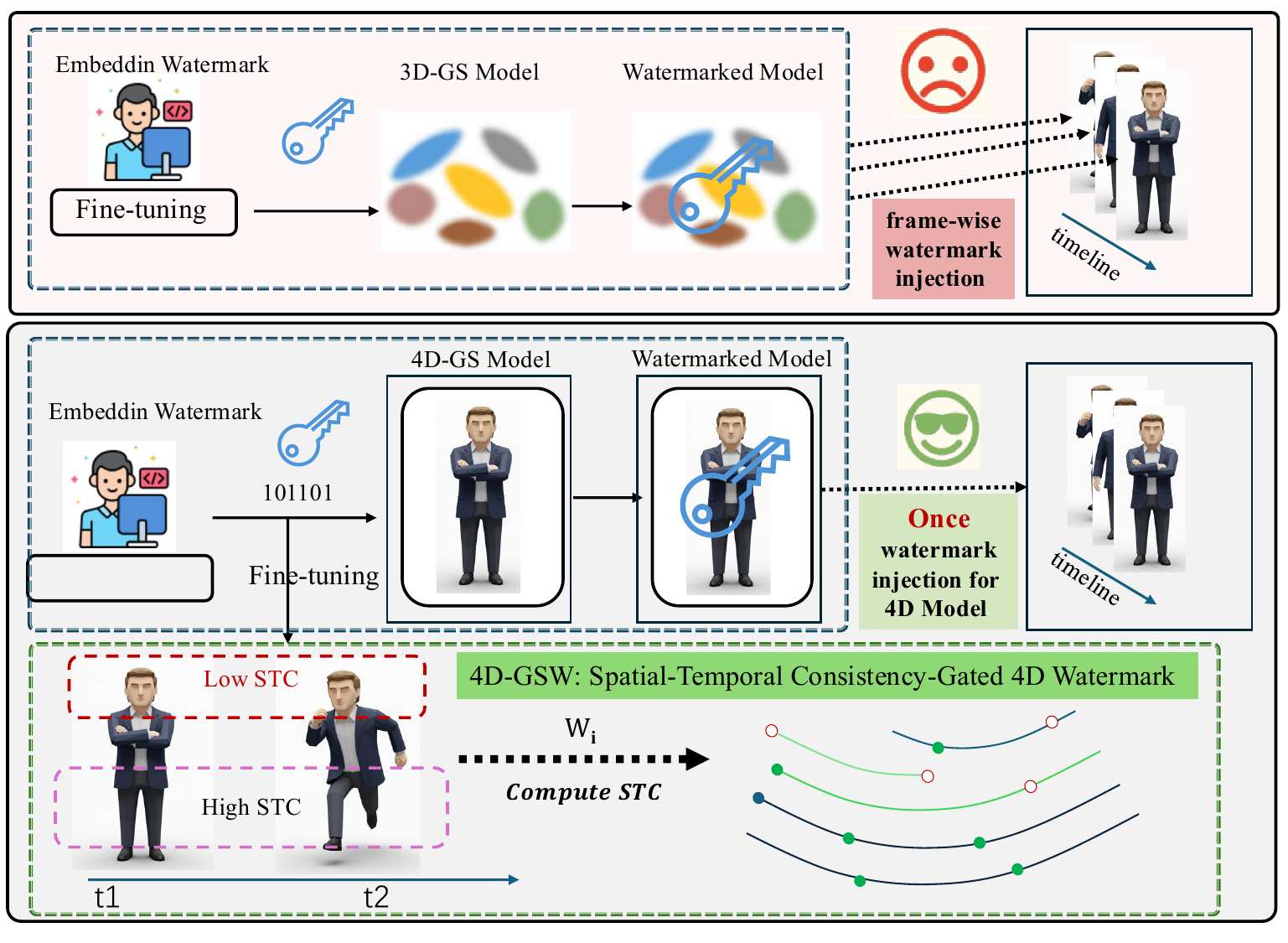}
\vspace{-3mm}
\caption{\textbf{Watermark framework comparison.}
 (\textbf{Upper}) Conventional 3D-GS watermarking schemes necessitate independent fine-tuning for each individual frame in a dynamic sequence. (\textbf{Bottom}) Our 4D-GSW treats the 4D asset as a holistic kinematic manifold. By computing the Spatio-Temporal Curvature (STC) of Gaussian trajectories, we adaptively gate the watermark gradients---prioritizing stable regions while shielding "Dynamic Instants".}
\vspace{-4mm}
\label{fig:compare}
\end{figure}

\begin{NoHyper}

Novel View Synthesis (NVS) is a crucial task in computer vision and robotics, with wide applications in autonomous driving and augmented reality. 3D Gaussian Splatting (3DGS)~\citep{kerbl3Dgaussians, Wang_2025_CVPR} has established itself as a prominent technique in 3D rendering, offering superior quality and fast rendering speeds. The 4D extensions of 3D-GS have demonstrated significant success in effectively reconstructing complex dynamic scenes\citep{Hu2025VTGSSLAM, chen2025haifgs}. The proliferation of 4D Gaussian Splatting (4D-GS)~\citep{Wu_2024_CVPR} has revolutionized the creation of high-fidelity dynamic digital assets, which now serve as cornerstones in autonomous driving simulation, virtual production, and telepresence~\citep{yan2024street, fang2024gaussianeditor, Jiang_2024_CVPR, zhang2025sqs}. However, the immense commercial value of these assets---derived from costly multi-view acquisitions and massive compute---is starkly contrasted by their inherent vulnerability to intellectual property (IP) infringement. Unlike implicit representations, the explicit parameterized nature of 4D-GS allows attackers to easily extract, redistribute, or maliciously manipulate motion trajectories and radiance attributes~\citep{yang2023gs4d}. Without robust and invisible copyright authentication, the large-scale deployment of these 4D assets remains high-risk, necessitating the development of specialized digital watermarking for dynamic neural rendering.

Despite the maturity of watermarking for 2D images and 3D meshes, extending these protections to 4DGS reveals a significant technical gap. A naive migration of static 3D-GS watermarking schemes to the temporal domain fails to account for the continuous spatio-temporal manifold structure of dynamic scenes~\citep{NEURIPS2024_59091e82, jang2024waterf, Chen_2025_CVPR, liu2025hide}. Specifically, injecting independent watermark perturbations into each frame disrupts the physical coherence of object trajectories, leading to "FVD collapse" characterized by severe temporal flickering and geometric jitter~\citep{unterthiner2019fvd, yin20234dgen, kplanes_2023}. While recent steganographic attempts like Hide-in-Motion~\citep{liu2025hide} achieve information hiding, they primarily focus on visibility and lack explicit constraints on the underlying motion kinematics, often sacrificing spatio-temporal consistency under complex non-rigid deformations~\citep{jiang2024consistentd, yin20234dgen}.

To bridge this gap, we propose \textbf{4D-GSW}, a kinematic-aware watermarking framework designed to reconcile the conflict between robust data embedding and 4D physical consistency. Our core design is rooted in the intrinsic manifold of motion: we observe that watermark-induced artifacts are most prominent at "Dynamic Instants" where trajectories exhibit high Spatio-Temporal Curvature (STC). By mathematically quantifying this curvature, we introduce an adaptive gating mechanism that shields these fragile motion segments from non-physical watermark gradients~\citep{9879461}. To ensure the persistent and coherent distribution of the watermark field, we formulate the embedding as a \textit{Maximum A Posteriori} (MAP) inference problem on a dynamic random field. Specifically, we integrate an \textit{Optimal Transport} (OT) alignment module to resolve the correspondence ambiguity caused by Gaussian densification and pruning, coupled with a joint HMM-MRF energy model for global coherence. We show that, in the continuum limit, the proposed objective can be interpreted as a curvature-modulated diffusion--reaction process, where the watermark signal naturally flows along stable radiance manifolds while being repelled by high-dynamic boundaries. Combined with an \textbf{anisotropic gradient routing} mechanism, our method ensures that high-bitrate copyright information is "baked" into the 4D asset with high-fidelity FVD degradation. The core contributions are summarized as follows:

\end{NoHyper}
\begin{itemize}
    \item \textbf{4D-GSW Framework:} We propose 4D-GSW, a kinematic-aware framework that bridges the gap between 4D asset protection and spatio-temporal consistency. To our knowledge, this is the first work to explicitly mitigate "FVD collapse" by aligning watermark embedding with 4D motion physics.
    
    \item \textbf{STC Gating Mechanism:} We introduce a \textbf{Spatio-Temporal Curvature (STC)} gating strategy that quantifies trajectory instability to adaptively shield "Dynamic Instants" from perturbations, preserving the integrity of complex non-rigid motions.

    \item \textbf{Topology-Aware Consistency:} We develop a joint \textbf{HMM-MRF} energy model integrated with Optimal Transport (OT). This resolves correspondence ambiguity during Gaussian resampling, enforcing rigorous temporal phase-locking and spatial coherence.

    \item \textbf{Continuum-Limit Interpretation:} We provide a continuum-limit analysis showing that the proposed curvature-gated objective can be interpreted as a curvature-modulated diffusion--reaction process, offering a mechanistic explanation for how watermark updates are redistributed over the spatio-temporal manifold.
    
\end{itemize}

\vspace{-2mm}
\section{Related Work}
\vspace{-2mm}
\paragraph{Dynamic Scene Rendering and Watermarking.} While 4D Gaussian Splatting (4DGS) and its sparse-controlled variants (e.g., SC4D~\citep{wu2024sc4d}) enable high-fidelity dynamic scene rendering, safeguarding their intellectual property remains an open challenge. Existing neural watermarking strategies primarily target static NeRFs or 3DGS~\citep{luo2023copyrnerf, Chen_2025_CVPR, guo2025splatssplatsrobusteffective, li2024gaussianmarker, NEURIPS2024_59091e82, zhang2025securegs}. Applying these static methods frame-by-frame to 4DGS invariably induces severe temporal discontinuity and structural instability. Furthermore, the only existing 4D steganography approach, Hide-in-Motion~\citep{liu2025hide}, relies on opacity heuristics without explicit modeling of the underlying motion physics, resulting in FVD degradation under complex non-rigid deformations. Since human perception is highly sensitive to abrupt motion changes---which can be mathematically localized via Spatio-Temporal Curvature (STC)~\citep{rao2002view}---naive spatial embedding is inadequate. Motivated by this, our work introduces kinematic-aware priors via STC to achieve temporally consistent and imperceptible 4D copyright embedding. More detailed related works are provided in Appendix~\ref{related}.

\vspace{-2mm}
\section{Preliminaries}
\vspace{-2mm}

\noindent\textbf{3D Gaussian Splatting.}
3DGS~\citep{kerbl3Dgaussians} represents a scene as anisotropic Gaussian
primitives parameterized by position, covariance, opacity, and
appearance coefficients, which are rendered through alpha compositing.
We use the zeroth-order SH coefficient as the watermark carrier.
Complete rendering notation is provided in Appendix~\ref{app:nomenclature}.

\noindent\textbf{4D Gaussian Splatting.}
Our watermarking framework is designed to be highly adaptive across 4D-GS backbones that model temporal dynamics\citep{luiten2023dynamic, Wu_2024_CVPR}. 4D-GSW requires temporally evolving Gaussian states, or
correspondences that can be estimated between adjacent timestamps,
to compute STC and spatio-temporal consistency.
Therefore, the framework is potentially compatible with other
temporal Gaussian representations satisfying these conditions. At any given timestamp $t$, the dynamic scene is represented by a set of appearance-augmented Gaussians $\mathcal{P}_t$:
\vspace{-2mm}
\begin{equation}
\begin{aligned}
    \mathcal{P}_t = \{ \mathbf{p}_i^t = (\mathbf{x}_i^t, \mathbf{q}_i^t, \alpha_i^t, \mathbf{s}_i^t) \}_{i=1}^N,
\end{aligned}
\end{equation}
where $\mathbf{x}_i^t$ and $\mathbf{q}_i^t$ denote the time-varying spatial coordinates and rotation state, respectively. To resolve naming conflicts with the geometric scaling factor $\boldsymbol{\sigma}$, we explicitly denote $\mathbf{s}_i^t$ as the \textbf{appearance state} (instantiated as the 0-th order Direct Current (DC) component of the SH coefficients). $\alpha_i^t$ denote the opacity. For notation simplicity in the subsequent optimization framework, we define $\mathbf{z}_i^t = [\mathbf{x}_i^t, \mathbf{s}_i^t]^\top$ as the coupled spatio-temporal state of the $i$-th Gaussian, representing its geometric center and appearance attribute, respectively. In our methodology, $\mathbf{s}_i^t$ serves as the primary carrier for watermark signal injection, as it directly influences the radiometric output while remaining decoupled from the kinematic deformation modeling.

\vspace{-3mm}
\subsection{Problem Formulation}
\label{subsec:background}
\vspace{-2mm}

The objective of our framework is to embed a robust, imperceptible, and omnidirectional watermark signal into a dynamic 4DGS asset~\citep{wu2024sc4d}. Unlike prior 4D steganography works that may rely on specific "check viewpoints" for information recovery, our task is to ensure that the watermark is recoverable from any arbitrary perspective in 360$^\circ$ space, effectively "baking" the copyright information into the 4D radiance field.

\noindent \textbf{Watermark Encoding and Embedding.} 
Let $\mathbf{B} \in \{0, 1\}^L$ be a binary watermark sequence of length $L$ (e.g., $L=48$). The embedding process is formulated as a strategic modulation of the appearance manifold $\mathcal{S} = \{ \mathbf{s}_i^t \}_{i,t}$ across the temporal sequence. We define an embedding function $\Psi$ that modifies the 0-th order SH coefficients:
\vspace{-2mm}
\begin{equation}
\begin{aligned}
\mathbf{\hat{s}}_i^t = \Psi(\mathbf{s}_i^t, \mathbf{B}, w_i^t),
\end{aligned}
\end{equation}
where $w_i^t$ is the kinematic-aware weight derived from the spatio-temporal curvature (STC). This ensures that the watermark signal evolves fluidly alongside the dynamic scene geometry $\mathbf{x}_i^t$ while strictly maintaining photometric fidelity.


\noindent \textbf{Omnidirectional 360$^\circ$ Supervision.} 
To enable watermark extraction from any viewing angle, we implement a dense multi-view supervision strategy during the optimization phase. We define a set of $M$ virtual viewpoints $\mathcal{V} = \{v_j\}_{j=1}^M$, sampled at uniform angular intervals of 15$^\circ$ around the asset. For each timestamp $t$ and viewpoint $v_j$, the rendered image $\hat{\mathcal{I}}^{t}_{v_j}$ must satisfy the recovery constraint:
\vspace{-2mm}
\begin{equation}
\begin{aligned}
\forall v_j \in \mathcal{V}, \quad \mathcal{D}(\hat{\mathcal{I}}^{t}_{v_j}) \approx \mathbf{B},
\end{aligned}
\end{equation}
where $\mathcal{D}$ is the pre-trained neural decoder\citep{10.1007/978-3-030-01267-0_40}. This formulation ensures that the copyright information is invariant from any observation point.

\vspace{-2mm}
\section{Methodology}
\label{method}
\vspace{-2mm}

\begin{figure}[!t]
\centering
\includegraphics[width=0.95\linewidth]{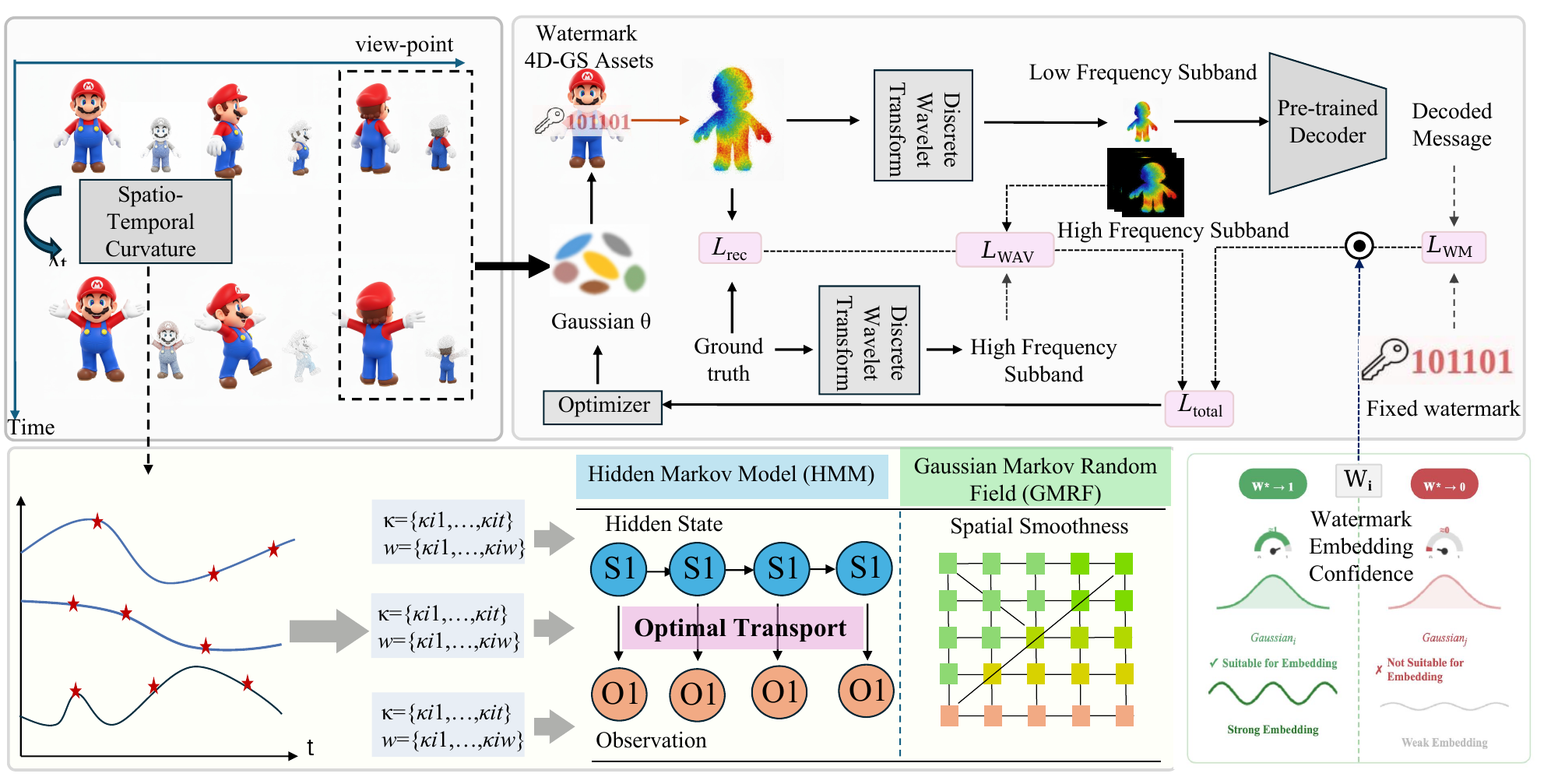}
\caption{\textbf{4D-GSW framework overview.}
Our pipeline extracts Spatio-Temporal Curvature (STC) from Gaussian trajectories to derive an embedding confidence weight. This weight guides a kinematic-aware optimization that adaptively modulates watermark gradients, prioritizing stable radiance manifolds while shielding dynamic instants. The field is further regularized by a joint HMM-GMRF energy model and wavelet-domain supervision to ensure high-fidelity 360$^\circ$ recovery.}
\vspace{-5mm}
\label{fig:framework}
\end{figure}

This section details the theoretical modeling and mathematical derivation of our curvature-modulated HMM-MRF spatio-temporal consistent watermarking framework for 4D-GS. As shown in Figure~\ref{fig:framework}, our method achieves covert and highly robust embedding of copyright information $\mathcal{B}$ within 4D space without compromising the original rendering capabilities of 4DGS.

\vspace{-2mm}
\subsection{Kinematic Analysis of Gaussian Trajectories
\label{curve}}
\vspace{-2mm}

In dynamic scene reconstruction, the parameter evolution of a 4D Gaussian field essentially describes the underlying motion manifold. The trajectory of a dense Gaussian $\mathbf{p}_i$ across continuous video frames can be modeled as a continuous function of time $\mathbf{x}_i(t)$. To quantify the dynamical instability of Gaussian trajectory, we define the instantaneous velocity $\mathbf{v}_i^t$ and acceleration $\mathbf{a}_i^t$:
\vspace{-2mm}
\begin{equation}
\begin{aligned}
\mathbf{v}_i^t = \frac{\mathbf{x}_i^t - \mathbf{x}_i^{t-1}}{\Delta t}, \quad \mathbf{a}_i^t = \frac{\mathbf{v}_i^t - \mathbf{v}_i^{t-1}}{\Delta t}.
\end{aligned}
\end{equation}
\vspace{-2mm}

According to classical curve geometry and motion perception theory, the instantaneous \textit{Spatio-Temporal Curvature} (STC) $\kappa_i^t$ ~\citep{rao2002view} measures the intensity of changes in the magnitude and direction of the velocity vector:
\vspace{-3mm}
\begin{equation}
\begin{aligned}
\kappa_i^t = \frac{\| \mathbf{v}_i^t \times \mathbf{a}_i^t \|_2}{\| \mathbf{v}_i^t \|_2^3 + \epsilon},
\end{aligned}
\end{equation}
where $\epsilon$ is a small constant for numerical stability. Regions with high $\kappa_i^t$ correspond to ``Dynamic Instants'' (e.g., sudden collisions or sharp turns), where the underlying parameter manifolds are highly sensitive and fragile. And these regions are visually salient and possess fragile underlying parameter manifolds that are susceptible to damage from watermark gradients. Injecting watermark gradients into such regions tends to distort the motion semantics, leading to severe visual artifacts. To mitigate this, we introduce a geometry-aware mapping that transforms STC into a \textit{Watermark Embedding Confidence Weight} $w_i^t$:
\vspace{-2mm}
\begin{equation}
\begin{aligned}
w_i^t = \exp\left(-\frac{\kappa_i^t}{\tau}\right)
\end{aligned}
\end{equation}
where $\tau$ is a decay factor. This weight acts as a gating signal for gradient routing: in stable regions (e.g., low-curvature points $w_i^t \to 1$), the system allows high-SNR watermark injection, whereas in high-dynamic regions (e.g., high-curvature points $w_i^t \to 0$), it enforces physical isolation for watermark injection to preserve motion integrity. To handle the potentially large dynamic range of $\kappa_i^t$ and prevent weight sparsity, we apply a mini-batch normalization to $w_i^t$ during training, followed by a sigmoid-like clamping to ensure the weights are well-distributed within the $[0, 1]$ interval.

\vspace{-3mm}
\subsection{Spatial-Temporal Consistency via Optimal Transport}
\label{Optimal_Transport}
\vspace{-2mm}

Since the watermark is extracted from 2D rendered images, the 4D Gaussian field must consistently carry watermark segments across frames to ensure robust decoding and eliminate high-frequency temporal oscillations (i.e., bit flicker). To achieve this temporal persistence while maintaining spatial coherence, we enforce a robust spatio-temporal consistency prior by modeling the evolution of the Gaussian field as a stochastic process on a dynamic random field. Standard 4D consistency constraints rely on fixed index-wise correspondences. However, during 4D-GS optimization, primitives undergo frequent densification, pruning, and resampling to maintain rendering quality, leading to topological mutations across frames. Such correspondence breakage introduces systematic bias into the consistency loss, manifesting as irregular oscillations during training.

To address this hidden-variable matching problem, we model the temporal evolution as a stochastic process. Specifically, we model the spatio-temporal states as a \textit{Hidden Markov Model (HMM)}, where the transition prior is established via \textit{Optimal Transport (OT)}~\citep{villani2009optimal,OT} to handle topological mutations. Furthermore, the spatial interactions are regularized through a \textit{Gaussian Markov Random Field (GMRF)} to ensure neighborhood coherence.

\noindent \textbf{Topological Alignment via Optimal Transport.} We resolve the aforementioned correspondence ambiguity by leveraging OT theory to infer the transition prior as a hidden-variable matching problem. We treat primitive sets at adjacent timestamps as discrete measures $\mu_t$ and $\mu_{t-1}$, and solve for the optimal transport plan $\gamma^*$ via the geometry-aware entropy-regularized Wasserstein distance~\citep{solomon2015convolutional}:
\vspace{-1mm}
\begin{equation}
\begin{aligned}
\min_{\gamma \in \Pi(\mu_t, \mu_{t-1})} \sum_{i,j} \gamma_{ij} \| \mathbf{x}_i^t - \mathbf{x}_j^{t-1} \|_2^2 - \epsilon H(\gamma).
\end{aligned}
\end{equation}
where $H(\gamma)$ is the entropic penalty solved efficiently via the Sinkhorn algorithm~\citep{cuturi2013sinkhorn}. To maintain training efficiency, the OT alignment is performed on a subsampled set of keyframes and Gaussian primitives. Based on the optimal plan $\gamma^*$, we employ a \textbf{barycentric mapping} to retrieve the aligned prior state $\hat{\mathbf{z}}_i^{t-1}$ for the $i$-th Gaussian:
\vspace{-3mm}
\begin{equation}
\begin{aligned}
\hat{\mathbf{z}}_i^{t-1} = \sum_{j} \frac{\gamma_{ij}^*}{\sum_k \gamma_{ik}^*} \mathbf{z}_j^{t-1},
\end{aligned}
\end{equation}
where $\mathbf{z} = [\mathbf{x}, \mathbf{s}]^\top$ encompasses both spatial coordinates and appearance attributes. This alignment ensures that consistency regularization operates on physical motion residuals $\Delta \mathbf{z}$ rather than noise.

\noindent \textbf{Gated Consistency Energy.} Using the OT-aligned reference $\hat{\mathbf{z}}_i^{t-1}$, the GMRF formulation unifies temporal transitions and spatial interactions into a \textbf{Gated Consistency Loss} $\mathcal{L}_{con}$:
\vspace{-2mm}
\begin{equation}
\begin{aligned}
\label{eq:con_loss}
\mathcal{L}_{con} = \sum_i w_i^t \left( \| \mathbf{z}_i^t - \hat{\mathbf{z}}_i^{t-1} \|_2^2 + \sum_{j \in \mathcal{N}i} \beta_{ij} \| \mathbf{z}_i^t - \mathbf{z}_j^t \|_2^2 \right).
\end{aligned}
\end{equation}

Here, $\mathcal{N}_i$ denotes the set of $K$-nearest neighbors for the $i$-th Gaussian, $\beta_{ij}$ denotes spatial adjacency weights. As established in Sec~\ref{curve}, the kinematic weight $w_i^t$ acts as a precision factor for this prior. It enforces a \textit{phase-locking} mechanism in stable regions to anchor watermark bits, while adaptively relaxing constraints at ``Dynamic Instants'' to ensure that regularization operates on true physical motion residuals rather than topological noise.

\vspace{-3mm}
\subsection{Curvature-Gated Spatial-Temporal MAP Optimization}
\vspace{-2mm}
To ensure the watermark signal is embedded both persistently and coherently, we formulate the optimization as a Maximum A Posteriori (MAP) inference problem. Within this framework, the search for the optimal Gaussian states $\mathcal{Z}_t$ is guided by the following energy functional:
\vspace{-2mm}
\begin{equation}
\label{eq:map_energy}
\begin{aligned}
\mathcal{E}(\mathcal{Z}t) = \mathcal{L}_{data} + \lambda_{con} \mathcal{L}_{con}(\mathcal{Z}t, \hat{\mathcal{Z}}{t-1}),
\end{aligned}
\end{equation}
where $\mathcal{L}_{data}$ represents the other joint reconstruction and watermark decoding loss (detailed in Sec.~\ref{loss}). The term $\mathcal{L}_{con}$, as defined in Eq.~\ref{eq:con_loss}, functions as a Spatial-Temporal Markov Prior. Crucially, the kinematic weight $w_i^t$ acts as a precision factor for this prior: it enforces a \textit{phase-locking} mechanism in dynamically stable regions ($w_i^t \to 1$) to anchor watermark bits, while adaptively relaxing constraints in ``Dynamic Instants'' ($w_i^t \to 0$). This probabilistic formulation ensures that the embedding process is not a mere perturbation but is regularized by the underlying motion manifold, providing the necessary boundary conditions for the anisotropic diffusion analyzed in Appendix~\ref{sec:theory_summary}.




\vspace{-2mm}
\subsection{Kinematic-Aware Joint Optimization}
\label{loss}
\vspace{-2mm}
To ensure that the embedded watermark remains robust and imperceptible within the dynamic 4D radiance field, we propose a \textbf{Kinematic-Aware joint optimization} objective.

\noindent \textbf{Frequency-Domain Watermark Supervision.}
To enhance the robustness of information hiding while maintaining high-frequency rendering fidelity, we perform watermark extraction and supervision in the frequency domain using Discrete Wavelet Transform (DWT). For a rendered image $\hat{\mathcal{I}}^{t}_{v_j}$ at timestamp $t$ from viewpoint $v_j \in \mathcal{V}$, the DWT operator $\Phi$ decomposes it into a low-frequency subband $\hat{\mathcal{I}}_{LL}$ and a set of high-frequency subbands $\hat{\mathcal{I}}_{H} = \{\hat{\mathcal{I}}_{LH}, \hat{\mathcal{I}}_{HL}, \hat{\mathcal{I}}_{HH}\}$ based on $[\hat{\mathcal{I}}{LL}, \hat{\mathcal{I}}{H}] = \Phi(\hat{\mathcal{I}}^{t}_{v_j})$.

\noindent\textbf{Kinematic-Aware Watermark Message Loss:} The $L$-bit watermark $\mathbf{B}$ is decoded exclusively from the stable $\hat{\mathcal{I}}_{LL}$ component. We employ a pre-trained message decoder $\mathcal{D}$ followed by a sigmoid function $\sigma(\cdot)$ to extract the message $\hat{\mathbf{B}} = \sigma(\mathcal{D}(\hat{\mathcal{I}}_{LL}))$ within the range of $[0, 1]$. The watermark message loss $\mathcal{L}_{wm}$ is defined as the Binary Cross Entropy (BCE) between $\hat{\mathbf{B}}$ and the GT $\mathbf{B}$:
\vspace{-2mm}
\begin{equation}
\begin{aligned}
\mathcal{L}_{wm} = -\sum_{i=1}^{N} (\mathbf{B}_{i} \log(\hat{\mathbf{B}}_{i}) + (1 - \mathbf{B}_{i}) \log(1 - \hat{\mathbf{B}}_{i})).
\end{aligned}
\end{equation} 
\noindent \textbf{Wavelet-subband Loss.} To prevent the watermark embedding degrading high-frequency structural details, we follow \citep{Jang_2025_CVPR} to adopt wavelet-subband loss $\mathcal{L}_{wav}$. This loss enforces consistency between the high-frequency coefficients of the watermarked image and the ground truth image $\mathcal{I}^{gt}$:
\vspace{-2mm}
\begin{equation}
\begin{aligned}
\mathcal{L}_{wav} = \sum_{l} \sum_{s \in \mathcal{H}} \mathbb{E} \left[ \left\| \Phi_{s,l}(\hat{\mathcal{I}}^{t}_{v_j}) - \Phi_{s,l}(\mathcal{I}^{gt}) \right\|_1 \right],
\end{aligned}
\end{equation}
where $\Phi_{s,l}(\cdot)$ is the DWT coefficients in subband $s$ at decomposition level $l$.

\noindent \textbf{Curvature-Modulated Joint Objective.} 
To spatially redistribute the watermark injection intensity, we reformulate the message loss $\mathcal{L}_{wm}$ as a kinetically-gating factor. By modulating the contribution of each Gaussian primitive to the message extraction gradient, we ensure that low-curvature points ($w_i^t \to 1$) act as high-SNR channels, while high-curvature points ($w_i^t \to 0$) suppress embedding. Finally, 4D-GSW is optimized with the total loss, which is the weighted sum of all losses:

\vspace{-3mm}
\begin{equation}
\begin{aligned}
\mathcal{L}_{total} = \lambda_{wm}\mathcal{L}_{wm} + \lambda_{wav}\mathcal{L}_{wav} + \lambda_{con}\mathcal{L}_{con} + \lambda_{rec}\mathcal{L}_{rec}
\end{aligned}
\end{equation}

where $\mathcal{L}_{rec} = \mathcal{L}_{ref} + \lambda_{l}\mathcal{L}_{lpips} + \lambda_{m}\mathcal{L}_{mask} + \lambda_{ga}\mathcal{L}_{ga} + \lambda_{sds}\mathcal{L}_{sds}$ aims to maintain the structural and photometric integrity of the 4D asset. Specifically, $\mathcal{L}_{ref}$ ensures photometric fidelity, $\mathcal{L}_{lpips}$ guarantees perceptual invisibility, $\mathcal{L}_{mask}$ enforces geometric boundaries, and $\mathcal{L}_{sds}$ ensures 3D-consistent geometry. $\mathcal{L}_{sds}$ ensures 3D-consistent geometry from monocular input. $\mathcal{L}_{ga}$ provides generative augmentation to further refine the detail synthesis in occluded regions. More loss details can refer to \citep{wu2024sc4d}. By concentrating the curvature-based gating $w_i^t$ exclusively on the watermarking terms, our framework induces an \textit{anisotropic gradient routing} mechanism. This allows the model to "bake" information into stable radiance manifolds without interfering with the complex motion optimization handled by the foundational losses, effectively preventing ``FVD collapse''.

\noindent \textbf{Kinematic-Aware Anisotropic Gradient Routing.} A naive optimization of Eq. (15) would uniformly distribute the watermark gradients across all Gaussian primitives, inevitably disrupting fragile motion trajectories and leading to "FVD collapse". To physically decouple the watermark embedding from complex scene dynamics, we implement a an \textit{anisotropic gradient routing} mechanism. 

Specifically, during backpropagation, the gradients of the watermark-related and consistency losses flowing into the appearance attribute (i.e., the 0-th order DC component $s_i^t$) of the $i$-th Gaussian are element-wise modulated by the kinematic weight $w_i^t$:
\vspace{-1mm}
\begin{equation}
\begin{aligned}
\frac{\partial \mathcal{L}_{total}}{\partial s_{i}^t} = \underbrace{\frac{\partial \mathcal{L}_{rec}}{\partial s_{i}^t}}_{\text{Reconstruction}} + w_{i}^{t} \cdot \underbrace{\frac{\partial (\lambda_{wm}\mathcal{L}_{wm} + \lambda_{wav}\mathcal{L}_{wav} + \lambda_{con}\mathcal{L}_{con})}{\partial s_{i}^t}}_{\text{Selective Embedding}}
\end{aligned}
\end{equation}
This asymmetric distribution of optimization pressure ensures that high-curvature primitives ($w_i^t \to 0$), which are critical for motion realism, naturally suppress watermark embedding. Conversely, dynamically stable primitives ($w_i^t \to 1$) receive strong watermark gradients. This gradient-level gating directly implements the anisotropic diffusion prior defined in Sec~\ref{Optimal_Transport}, isolating watermark-induced perturbations from non-rigid deformations.
\vspace{-3mm}
\section{Experiments}
\label{exp}
\vspace{-2mm}

\noindent\textbf{Dataset and Metrics.} We evaluate on the Consistent4D dataset~\citep{jiang2024consistentd}, utilizing seven animated assets for quantitative analysis and real-world monocular videos for qualitative validation. Following standard protocols, we reconstruct scenes from a monocular view and evaluate Novel View Synthesis (NVS) on four unseen views. To comprehensively assess the 4D assets~\citep{wu2024sc4d,guo2025splatssplatsrobusteffective,liu2025hide,li2024gaussianmarker}, we report PSNR and SSIM~\citep{wang2004ssim} (photometric fidelity), LPIPS~\citep{zhang2018perceptual} (perceptual quality), CLIP score~\citep{radford2021clip} (semantic consistency), and Fréchet Video Distance (FVD)~\citep{unterthiner2019fvd} (temporal realism) to quantify the distribution shift in the I3D~\citep{carreira2017quo} feature space.

\noindent\textbf{Implementation Details.} We instantiate 4D-GSW on SC4D~\citep{wu2024sc4d}, which is used for all
quantitative experiments. STC-guided watermark optimization is activated
after the reconstruction warm-up stage. Since the official HiM
implementation is unavailable, we independently reimplement it under
matched experimental settings. 4D-GSW adds approximately $25\%$
training time over SC4D without additional rendering-time overhead.
Full implementation and reproduction details are provided in
Appendix~\ref{app:implementation}. Additional trajectory-noise sensitivity results are provided in Appendix~\ref{app:trajectory_noise}.

\vspace{-2mm}
\subsection{Quantitative Comparison}
\label{quantitative_results}
\vspace{-1mm}

\begin{table}[h]
\vspace{-3mm}
\centering
\small
\caption{Quantitative comparison on dynamic scene reconstruction and watermarking. (* Reproduced results.)}
\resizebox{\linewidth}{!}{
\begin{tabular}{lccccc}
\toprule[1pt] 
    Methods & PSNR~$\uparrow$ & SSIM~$\uparrow$ & LPIPS~$\downarrow$ & CLIP~$\uparrow$ & FVD~$\downarrow$   \\
    \midrule 
    Consistent4D~\citep{jiang2024consistentd} & 27.62 & 0.89 & 0.132 & 0.908 & 1518.56  \\
    4DGen~\citep{yin20234dgen} & 21.80 & 0.90 & 0.130 & 0.894 &  -  \\
    4Diffusion~\citep{zhang2024diffusion} & - & - & 0.228 & 0.873 &  1551.63  \\
    L4GM~\citep{ren2024lgm} & 21.80 & 0.90 & 0.158 & \textbf{0.913} &  1360.04  \\ \midrule
    HiM*~\citep{liu2025hide} & 23.24 & 0.83 & 0.226 & 0.766 & 2095.54 \\
    SC4D~\citep{wu2024sc4d} & \textbf{29.50} & \textbf{0.95} & \textbf{0.080} & 0.900 & \textbf{1297.18} \\
    SC4D~\citep{wu2024sc4d} + HiDDeN~\citep{10.1007/978-3-030-01267-0_40} & 27.08 & 0.89 & 0.155 & 0.880 & 1601.50   \\ 
    SC4D~\citep{wu2024sc4d} + 3D-GSW~\citep{Jang_2025_CVPR} & 23.31 & 0.84 & 0.206 & 0.815 & 1820.50   \\ \midrule
    \rowcolor{colLoRA}
    \textbf{4D-GSW (Ours)} & 28.36 & 0.91& 0.133 & 0.904 & 1331.19  \\
\bottomrule[1pt] 
\end{tabular} 
}
\vspace{-3mm}
\label{tab:quant}
\end{table}

\noindent\textbf{Reconstruction Fidelity and Temporal Consistency.} 
As shown in Table~\ref{tab:quant}, \textbf{4D-GSW} consistently outperforms the watermarking baseline across most metrics. Detailed per-sample results refer to Table~\ref{tab:visual_quality_comp} in Appendix. \textbf{4D-GSW} outperforms the watermarking baselines across most metrics, achieving an FVD of 1331.19 compared with 1601.50 for SC4D+HiDDeN and 1820.50 for SC4D+3D-GSW.  Under the matched optimization budget, watermark embedding introduces only a small fidelity cost relative to non-watermarked SC4D. Detailed per-sample results are in Table~\ref{tab:visual_quality_comp}.


\begin{table*}[h]
\vspace{-2mm}
\caption{Robustness evaluation against various distortions. We report the bit accuracy of the extracted watermark (a 48-bit string) across different attack scenarios. (* Reproduced results by us)}
\centering
\footnotesize
\setlength{\tabcolsep}{0.95mm}
\label{table:robustness}
\resizebox{0.99\textwidth}{!}{
\begin{tabular}{lccccccccc}
\toprule
\multirow{2}{*}{Methods}               &\multirow{2}{*}{None} &Noise   &Rotation   &Scaling   &Blur      &Crop   &Resize  &JPEG    &\multirow{2}{*}{Average}\\
&           &($\mu$=0.1)&($\pm$$\pi$/6)&($\leq$25\%)&($\sigma$=$0.1$)&(20\%)  &(0.9$\sim$1.0)&(50\% quality)         &             \\
\midrule
HiM*~\citep{liu2025hide}  & 89.54           &  81.92       & 77.39 &  88.98  &  89.38  &   87.76   &  89.42 &  88.24                    &      86.58    \\[1ex]
SC4D~\citep{wu2024sc4d} + HiDDeN~\citep{10.1007/978-3-030-01267-0_40}       & 97.33           &  94.01       & 94.27 &   94.14  &  94.60  &   94.92   &  95.44 &  93.23                    &      94.74    \\[1ex]
SC4D~\citep{wu2024sc4d} + 3D-GSW~\citep{Jang_2025_CVPR}       & 97.28           &  94.85       & 82.18 &   97.40  &  97.25  &   96.36   &  97.37 &  97.35                    &      95.01    \\
\midrule
\rowcolor{colLoRA}
\textbf{4D-GSW (Ours)}  &\textbf{99.10}      &\textbf{96.84} &\textbf{98.65}  &\textbf{97.97} &\textbf{97.94}   &\textbf{98.00}&\textbf{97.98}  &\textbf{97.89}           &\textbf{98.05}        \\
\bottomrule[1pt]
\end{tabular}
}
\vspace{-2mm}
\end{table*}

\noindent\textbf{Robustness against Visual Distortions.}
As shown in Table~\ref{table:robustness}, \textbf{4D-GSW} achieves
\textbf{98.05\%} average bit accuracy and consistently outperforms all
compared watermarking baselines. It reaches \textbf{99.10\%} in the
clean setting and remains above \textbf{96.8\%} under all evaluated
image-space distortions, benefiting from our \textit{Topology-Aware Consistency Model} which stabilizes the watermark signal within the 4D radiance field. These results validate that our \textit{anisotropic diffusion} creates a redundant, robust representation across stable manifolds that is highly resistant to both local and global distortions.


\begin{table}[h]
\vspace{-2mm}
    \centering
    \caption{Asset-level robustness evaluation under direct Gaussian-space attacks. 
    The watermark is extracted after manipulating the trained 4D Gaussian asset.}
    \label{tab:asset_robustness}
    \resizebox{\linewidth}{!}{
    \begin{tabular}{lllc}
        \toprule
        \textbf{Attack Category} & \textbf{Attack Type} & 
        \textbf{Attack Parameter} & \textbf{Bit Accuracy (\%) $\uparrow$} \\
        \midrule
        None 
        & Clean Watermarked Asset 
        & No attack 
        & 99.10 \\
        \midrule
        Geometry Perturbation 
        & Gaussian Rotation 
        & $\theta = 45^\circ$ 
        & 93.80 \\
        
        \midrule
        Appearance Perturbation 
        & Opacity Noise 
        & $\sigma = 1.0$ 
        & 95.30 \\
        
        Appearance Perturbation 
        & Scale Noise 
        & $\sigma = 0.5$ 
        & 92.00 \\
        
        \midrule
        Gaussian Attribute Perturbation 
        & Covariance Rotation Noise 
        & $\sigma = 0.5$ 
        & 94.40 \\
        
        \midrule
        Gaussian Pruning 
        & Opacity-based Pruning 
        & 50\% primitives removed 
        & 95.00 \\
        
        \midrule
        Control Motion Perturbation 
        & Control XYZ Noise 
        & $\sigma = 0.1$ 
        & 93.10 \\
        
        Temporal Perturbation 
        & Time Deformation Noise 
        & $\sigma = 0.02$ 
        & 93.00 \\
        \bottomrule
    \end{tabular}
    }
\vspace{-5mm}
\end{table}

\noindent\textbf{Asset-level Robustness.}
We evaluate four types of direct attacks on the trained 4D Gaussian asset: geometric perturbations, appearance perturbations to opacity and scale, Gaussian pruning, and temporal motion perturbations. As shown in Table~\ref{tab:asset_robustness}, bit accuracy decreases only moderately from 99.10\% in the clean setting and remains above 95\% even after pruning 50\% of the Gaussian primitives. These results indicate that the watermark is distributed across the Gaussian representation and remains recoverable under substantial modifications to its geometry, and temporal dynamics.

\begin{wraptable}{r}{0.62\linewidth}
\vspace{-12pt}
\caption{Ablation study of different components in 4D-GSW.}
\label{tab:ablation}
    \vspace{-2mm}
    \begin{adjustbox}{max width=\linewidth}
        \begin{tabular}{ccccccc}
\toprule
\toprule[1pt] 
    Methods & PSNR~$\uparrow$ & SSIM~$\uparrow$ & LPIPS~$\downarrow$ & CLIP~$\uparrow$ & FVD~$\downarrow$ & ACC~$\uparrow$  \\
    \midrule 
    w/o temporal & 26.34 & 0.87 & 0.16 & 0.85 & 1825.70 & 94.60 \\
    w/o spatial & 25.21 & 0.88 & 0.17 & 0.87 &  1688.47  & 96.84\\
    w/o curvature & 26.12 & 0.90 & 0.15 & 0.88 &  1491.14 & 96.21\\
    \midrule
    \rowcolor{colLoRA}
    \textbf{4D-GSW (Ours)} & \textbf{28.36} & \textbf{0.91} & \textbf{0.133} & \textbf{0.904} & \textbf{1331.19}  & \textbf{99.10}\\
\bottomrule[1pt] 
\end{tabular}  
    \end{adjustbox}
    \vspace{-4mm}
\end{wraptable}
\noindent\textbf{Ablation Study.} Table~\ref{tab:ablation} validates the contribution of our core components. 
\textit{1) Temporal Consistency:} Removing the OT-aligned HMM prior (\textbf{w/o temporal}) causes severe ``FVD collapse'' (1825.70) and drops ACC to 94.60\%, proving that temporal phase-locking is essential for both motion realism and decoding robustness. 
\textit{2) Spatial Regularization:} Disabling the GMRF prior (\textbf{w/o spatial}) degrades PSNR to 25.21 dB, confirming that spatial affinity is critical for smooth signal distribution along the radiance manifold. 
\textit{3) Curvature Gating:} Excluding STC-based gradient routing (\textbf{w/o curvature}) spikes FVD to 1491.14, demonstrating the necessity of shielding dynamic instants to prevent trajectory distortion. 
Overall, the full \textbf{4D-GSW} achieves the favorable fidelity-robustness trade-off, confirming that our anisotropic optimization decouples watermark embedding from complex 4D scene geometry.

\vspace{-2mm}
\subsection{Qualitative Results}
\vspace{-2mm}
\begin{figure*}[h]
\vspace{-2mm}
    \centering
    \includegraphics[width=0.9\textwidth]{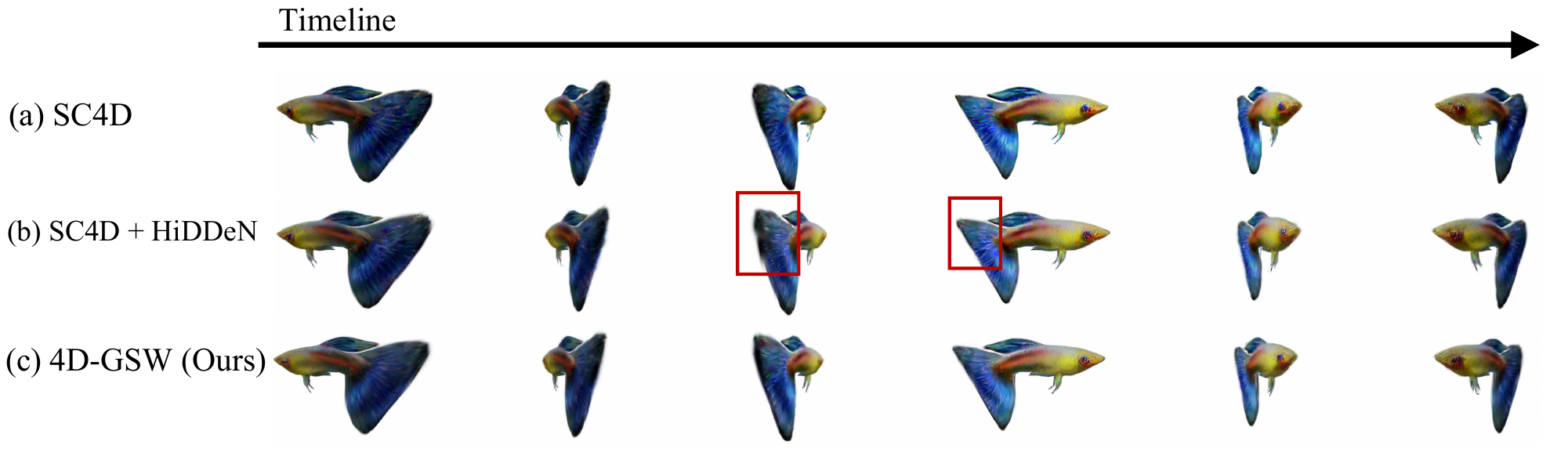}
    \vspace{-3mm}
    \caption{\textbf{Visualization of temporal consistency on the \textit{Guppie} sequence.} The baseline (b) suffers from severe temporal flickering and geometric jitter during rapid motion (red boxes). In contrast, our \textbf{4D-GSW} (c) maintains high structural integrity and motion smoothness consistent with the baseline.}
  \label{fig:timeline}
  \vspace{-2mm}
\end{figure*}

\begin{figure*}[htbp]
    \vspace{-2mm}
    \centering
    \includegraphics[width=0.99\textwidth]{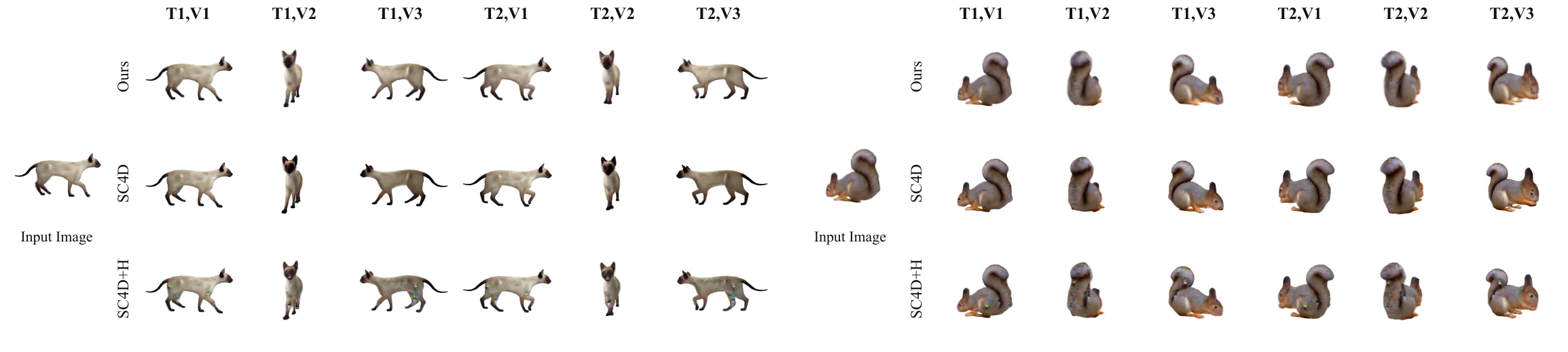}
    \caption{\textbf{Multi-view 4D Generation Results at Different Timestamps.} 
  Comparison of models generated by SC4D~\citep{wu2024sc4d}, SC4D+Hidden~\citep{10.1007/978-3-030-01267-0_40}, and our method across different viewpoints at the different timestep.}
  \label{fig:Multi-perspective}
  \vspace{-6mm}
\end{figure*}

\begin{figure*}[!h]
    \centering
    \includegraphics[width=0.85\textwidth]{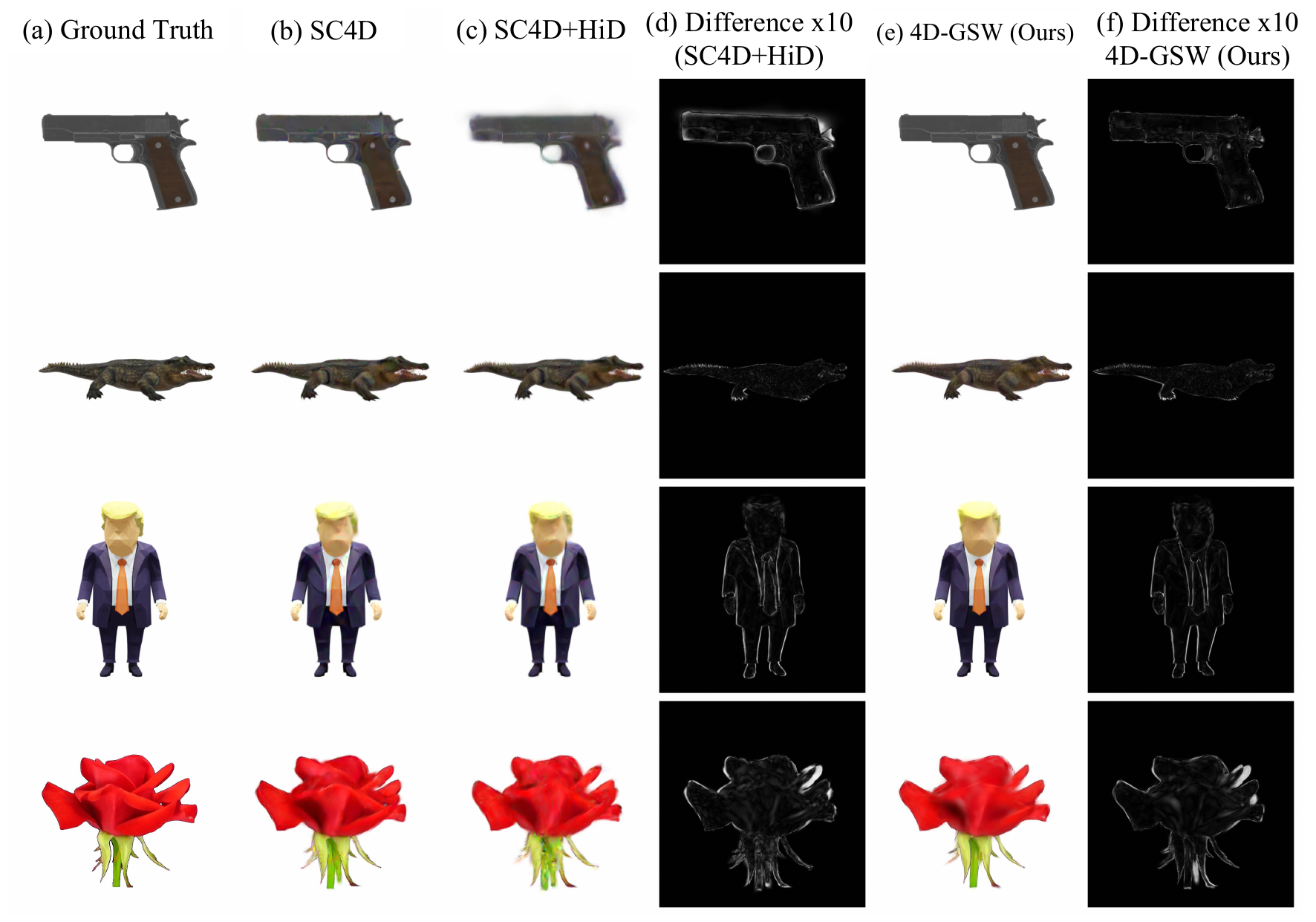}
    \caption{\textbf{Visualization across diverse dynamic assets.} 
  Comparison of models generated by SC4D, SC4D+Hidden, and our method across different viewpoints at the same time step.}
  \label{fig:Multi-view}
  \vspace{-5mm}
\end{figure*}

\textit{1) Spatio-Temporal and Cross-View Consistency:}
As shown in Fig.~\ref{fig:timeline} and Fig.~\ref{fig:Multi-perspective},
\textbf{SC4D+HiDDeN} exhibits temporal flickering and geometric jitter,
whereas \textbf{4D-GSW} preserves coherent dynamics and stable geometry
across timestamps and viewpoints.

\textit{2) Photometric Fidelity and Anisotropic Error Distribution:}
Fig.~\ref{fig:Multi-view} shows that \textbf{4D-GSW} preserves sharper
structural details and yields smaller residuals than the baseline,
particularly around fine structures. More visualizations are provided
in Appendix~\ref{more_visualization}.


\vspace{-5mm}
\section{Conclusion}
\label{conclusion}
\vspace{-3mm}

In this paper, we presented \textbf{4D-GSW}, a kinematic-aware watermarking framework for 4D Gaussian Splatting. To address the temporal artifacts induced by naive frame-wise embedding, we introduce a Spatio-Temporal Curvature (STC) metric that adaptively gates watermark gradients, shielding sensitive dynamic instants from non-physical perturbations. By formulating the embedding process as a joint HMM-MRF optimization regularized by Optimal Transport, our method successfully handles topological mutations to maintain rigorous spatio-temporal consistency. Rather than arbitrary injection, this curvature-gated mechanism acts as an anisotropic structural regularizer, effectively embedding robust copyright information into stable radiance manifolds. Extensive evaluations demonstrate that 4D-GSW achieves high resilience against diverse distortions with minimal degradation to motion realism, providing a practical foundation for IP protection in dynamic neural rendering.



\noindent \textbf{Discussion and Limitations.} Our STC metric assumes $C^2$ continuous Gaussian trajectories. During extreme non-rigid events (e.g., topological fracturing), this continuity is violated, rendering the kinematic manifold non-differentiable. Consequently, localized oscillations in the embedding weight $w_i^t$ can weaken the diffusion barrier, marginally degrading watermark robustness in these singular regions. Addressing such stochastic dynamics remains an avenue for future work.

\subsection*{AI use statement}
In this work, we used AI tools to check the typos in the paper. We have reviewed all AI-assisted work and take responsibility for the final content of this work.

\bibliography{refs}
\bibliographystyle{iclr2027_conference}

\appendix
\appendix

\newpage
\raggedbottom

\section{More Visualization Results}
\label{more_visualization}
\begin{figure*}[t]
    \centering
    \includegraphics[width=\textwidth]{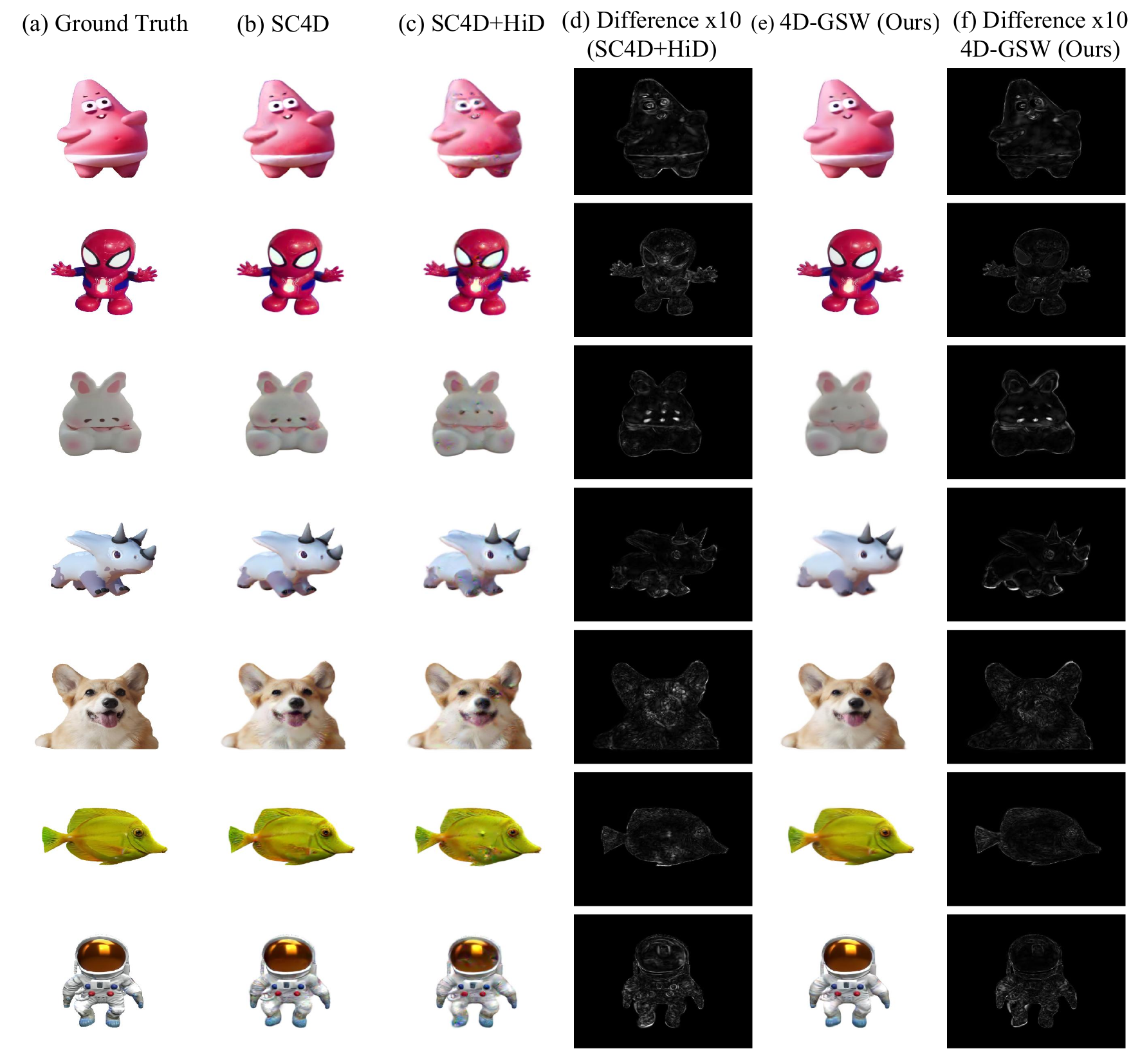}
    \vspace{-3mm}
    \caption{\textbf{Visualization across diverse dynamic assets.} 
  From left to right: (a) Ground Truth, (b) original SC4D, (c, d) SC4D+HiDDeN baseline and its $10\times$ residual map, (e, f) our \textbf{4D-GSW} and its $10\times$ residual map. While the baseline introduces blurring at structural boundaries, 4D-GSW achieves near-lossless reconstruction.}
  \label{fig:comparison}
  \vspace{-5mm}
\end{figure*}

\begin{figure*}[h]
    \centering
    \includegraphics[width=0.99\textwidth]{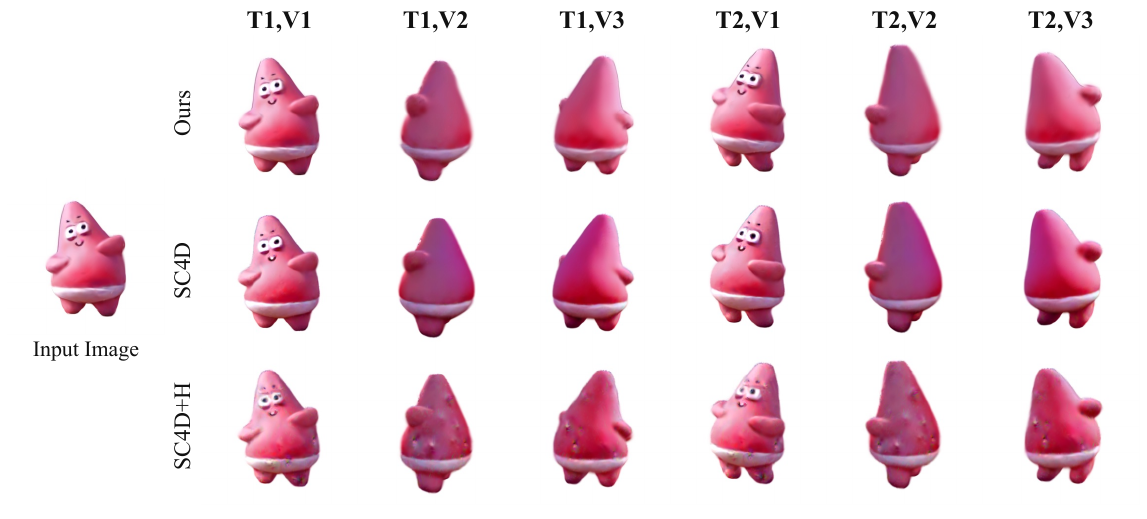}
    \caption{\textbf{Multi-view 4D generation at different timestamps.} 
  Qualitative comparison of SC4D, SC4D+Hidden, and our method under varying viewpoints at a fixed time step.}
  \label{fig:Multi-perspective2}
  \vspace{-5mm}
\end{figure*}

Fig.~\ref{fig:comparison} and Fig.~\ref{fig:Multi-perspective2} provide a detailed view of reconstruction quality and residual distributions. Compared to the baseline which introduces blurring at structural boundaries, \textbf{4D-GSW} preserves sharp textures and fine-grained details consistent with the \textbf{Ground Truth}.

\section{Related Work}
\label{related}
\paragraph{Dynamic 3D Scene Representations and 4DGS.} While dynamic Neural Radiance Fields (NeRFs)~\citep{barron2021mip, barron2023zipnerf, mildenhall2020nerf, pumarola2020d, park2021nerfies} achieve high-fidelity novel view synthesis via deformation networks, their immense computational cost limits practical deployment. To address this, 3D Gaussian Splatting (3DGS)~\citep{kerbl3Dgaussians} enables real-time rendering via explicit point-based representations. Its temporal extensions (4DGS)~\citep{bae2024ed3dgs, yang2023deformable3dgs, liu2025foundation} further model dynamic scenes. However, dense trajectory prediction per frame often induces optimization artifacts~\citep{Wu_2024_CVPR}. Alternatively, the Sparse-Controlled 4DGS (SC4D) framework~\citep{wu2024sc4d} decouples appearance and motion by driving dense Gaussians with a minimal set of sparse control points via Linear Blend Skinning (LBS), achieving superior local rigidity and spatio-temporal fidelity. Despite these reconstructive advances, the intellectual property (IP) protection of 4DGS assets remains largely unexplored. A core challenge is ensuring watermark imperceptibility across complex spatio-temporal domains. Human visual perception is highly sensitive to abrupt motion discontinuities, which can be mathematically localized using Spatio-Temporal Curvature (STC)~\citep{rao2002view}---a metric characterizing the misalignment between velocity and acceleration. Since perturbations in high-STC regions easily trigger visible artifacts, naive frame-wise watermark injection is inadequate. Motivated by this, we build upon the efficient SC4D baseline and introduce motion-aware priors, utilizing STC to guide robust and imperceptible 4D copyright embedding.

\vspace{-2mm}
\paragraph{Digital Watermarking in Neural Rendering.}
Neural rendering watermarking has evolved from hiding data in implicit NeRF weights~\citep{luo2023copyrnerf, Li_2023_ICCV} to more sophisticated strategies for explicit 3DGS representations. Existing 3DGS schemes primarily focus on: (i) modulating Spherical Harmonic (SH) coefficients with cross-modal priors or encryption~\citep{Chen_2025_CVPR, guo2025splatssplatsrobusteffective}; (ii) strategic densification in high-uncertainty or high-frequency regions~\citep{li2024gaussianmarker, Jang_2025_CVPR} ; and (iii) incorporating 3D distortion layers during training for geometric robustness~\citep{tan2024watergscopyrightprotection3d}. However, applying these static methods frame-by-frame to 4DGS induces severe temporal discontinuity. While Hide-in-Motion~\citep{liu2025hide} introduces 4D steganography via composite attributes, it relies on opacity heuristics and lacks explicit modeling of motion kinematics. This neglect of the underlying physical manifold leads to significant FVD degradation under complex non-rigid deformations. Furthermore, recent advancements in 3D Gaussian Splatting (3DGS) steganography and protection, such as GS-Hider~\citep{NEURIPS2024_59091e82} and SecureGS~\citep{zhang2025securegs}, have demonstrated the ability to hide large amounts of data within static scenes. Splats in Splats~\citep{guo2025splatssplatsrobusteffective} introduced a nested structure for embedding scenes within scenes. Although these methods represent state-of-the-art performance in the 3D domain, they lack modeling of the temporal dimension. This makes it difficult to ensure temporal consistency, often leading to flickering artifacts and structural instability, and thus they cannot be directly applied to the processing of 4D assets.

\section{3D Gaussian Splatting.}
\label{app:nomenclature}

3D Gaussian Splatting (3D-GS)~\citep{kerbl3Dgaussians} represents a static scene as a set of $N$ anisotropic Gaussian ellipsoids $\mathcal{G}=\{\mathbf{g}_i\}_{i=1}^{N}$. Each Gaussian is defined by its center position (mean) $\boldsymbol{\mu}_i \in \mathbb{R}^3$, an opacity $\alpha_i \in [0,1]$, and a 3D covariance matrix $\boldsymbol{\Sigma}_i$. Generally, $\boldsymbol{\Sigma}_i$ is decomposed into a scaling factor $\boldsymbol{\sigma}_i \in \mathbb{R}^3$ and a rotation quaternion $\mathbf{q}_i \in \mathbb{R}^4$. Besides, view-dependent color is encoded via spherical harmonic (SH) coefficients $\mathbf{h}_i$. During rendering, the pixel color $\hat{\mathbf{C}}(\mathbf{r})$ is computed through alpha compositing:
\vspace{-2mm}
\begin{equation}
\label{eq:render}
\begin{aligned}
  \hat{\mathbf{C}}(\mathbf{r})
    = \sum_{k=1}^{K} \mathbf{c}_k \,\alpha_k \,G_k'(\mathbf{r})
      \prod_{l=1}^{k-1}\bigl(1-\alpha_l\,G_l'(\mathbf{r})\bigr),
\end{aligned}
\end{equation}
where $\mathbf{c}_k$ is the color decoded from $\mathbf{h}_k$ and $G_k'(\mathbf{r})$ is the projected 2D Gaussian influence at pixel $\mathbf{r}$.

\section{Implementation Details and Computational Analysis}
\label{app:implementation}

\subsection{More Implementation Details}

We instantiate our framework on the Sparse-Controlled 4D-GS (SC4D) backbone~\citep{wu2024sc4d}, where $N$ dense Gaussians are driven by a sparse set of control points ($M \approx 512$) via Linear Blend Skinning (LBS). All quantitative experiments in this work are conducted using SC4D. SC4D represents the dynamic asset using dense Gaussian primitives driven by approximately 512 sparse control points through Linear Blend Skinning. The principal experimental settings are summarized in Table~\ref{tab:implementation_settings}.  The proposed STC gating, OT alignment, and rendered-view watermark supervision are not coupled to SC4D-specific sparse control points. While SC4D ensures local geometric rigidity, our kinematic-aware constraints specifically regularize the temporal smoothness of the appearance carrier $\mathbf{s}_i^t$ against watermark-induced high-frequency noise. Following SC4D's two-stage optimization strategy, we compute STC only after the reconstruction warm-up stage, when Gaussian trajectories have become sufficiently stable, thereby avoiding unreliable gradient routing caused by early-stage optimization noise. For topology changes caused by densification, pruning, or resampling, our OT-based Gaussian alignment establishes soft correspondences across timestamps before temporal consistency computation. Moreover, since STC only weights watermark-related gradients and does not directly constrain the reconstruction dynamics, imperfect STC estimation affects only the watermark embedding strength rather than destabilizing the learned Gaussian motion. Our message decoder architecture is consistent with that of HiDDeN~\citep{10.1007/978-3-030-01267-0_40} and equipped with fine-tuned pre-trained weights. We implemented our framework in PyTorch, training each scene for 10,000 iterations ($\sim$1 hour) on a single NVIDIA RTX A6000 GPU. Since the official implementation of HiM~\citep{liu2025hide} is unavailable, we reimplemented it based on the algorithmic description in the paper.  Besides, we matched the SC4D backbone, training budget, message length, rendering views, and attack settings whenever applicable. 


\begin{table}[h]
    \centering
    \caption{Implementation and evaluation settings of 4D-GSW.}
    \label{tab:implementation_settings}
    \small
    \setlength{\tabcolsep}{6pt}
    \begin{tabular}{ll}
        \toprule
        \textbf{Item} & \textbf{Setting} \\
        \midrule
        Backbone & SC4D \\
        Sequence length & 32 frames \\
        Watermark length & 48 bits \\
        Novel viewpoints per iteration & 4 \\
        Sparse control points & $\sim$512 \\
        Final Gaussian count & 22K--84K \\
        Message decoder & Fine-tuned HiDDeN-style decoder \\
        Hardware & NVIDIA RTX A6000, 48\,GB \\
        \bottomrule
    \end{tabular}
\end{table}

\subsection{Computational Overhead and Scalability}
Compared with the original SC4D optimization, 4D-GSW introduces
additional training-time components, including STC computation,
OT-based Gaussian alignment, spatio-temporal consistency regularization,
and multi-view watermark supervision. We measure the per-scene training
time on a single NVIDIA RTX A6000 GPU with 48\,GB memory under the same
SC4D reconstruction setting. 4D-GSW increases the training
time from approximately 60 minutes to 75 minutes per scene, corresponding
to a moderate overhead of approximately 25\%.

Importantly, these additional components are used only during
optimization and do not modify the original 4DGS rendering pipeline.
Therefore, the generated watermarked asset introduces no additional
rendering-time overhead. The additional training cost mainly comes from
OT alignment and the extra rendering--decoding passes required by
multi-view watermark supervision. In contrast, STC computation requires
only lightweight velocity and acceleration statistics and contributes
marginal overhead.

Regarding scalability, the computational cost of multi-view watermark
supervision grows approximately linearly with the number of sampled
viewpoints, since each additional viewpoint requires one additional
rendering and decoding pass. In our implementation, four novel
viewpoints are sampled per iteration. For OT alignment, we avoid
constructing a full transport matrix over all Gaussian primitives by
performing correspondence estimation on sampled Gaussian subsets and
selected temporal states. Therefore, the practical OT cost depends on
the sampled subset size rather than the full Gaussian count. Across the
evaluated scenes, the final number of Gaussian primitives ranges from
approximately 22K to 84K, and all sequences contain 32 frames.

\section{Detailed Quantitative Results}
\label{per_sample}
Refer to Table~\ref{tab:visual_quality_comp} and Table~\ref{tab:watermark_robustness_comp}, we also provide the per-sample results for visual quality and watermark robustness evaluation.

\begin{table}[t]
\centering
\small 
\caption{Per-sample visual quality metrics comparison. Higher PSNR, SSIM, CLIP scores and lower LPIPS, FVD indicate better visual fidelity.}
\label{tab:visual_quality_comp}
\resizebox{\linewidth}{!}{
\begin{tabular}{llccccc}
\toprule
Object & Method & PSNR $\uparrow$ & SSIM $\uparrow$ & LPIPS $\downarrow$ & CLIP $\uparrow$ & FVD $\downarrow$ \\
\midrule
\multirow{2}{*}{Aurorus} & SC4D\citep{wu2024sc4d} + 3D-GSW\citep{Jang_2025_CVPR} & 24.370 & 0.8804 & 0.1824 & 0.7322 & 2226.2  \\
                           & \textbf{4D-GSW} (Ours)  & 26.683 & 0.9241 & 0.1308 & 0.8382 & 1948.23 \\
\midrule
\multirow{2}{*}{Crocodile} & SC4D\citep{wu2024sc4d} + 3D-GSW\citep{Jang_2025_CVPR} & 24.815 & 0.8886 & 0.1311 & 0.8544 & 1611.9 \\
                           & \textbf{4D-GSW} (Ours) & 26.423 & 0.9038 & 0.1116 & 0.8598 & 1688.47 \\
\midrule
\multirow{2}{*}{Guppie} 
                           & SC4D\citep{wu2024sc4d} + 3D-GSW\citep{Jang_2025_CVPR} & 22.538 & 0.8573 & 0.1630 & 0.8735 & 1014.5 \\
                           & \textbf{4D-GSW} (Ours) & 26.347 & 0.8978 & 0.1086 & 0.9353 & 764.11 \\
\midrule
\multirow{2}{*}{Monster} 
                           & SC4D\citep{wu2024sc4d} + 3D-GSW\citep{Jang_2025_CVPR} & 24.344 & 0.8066 & 0.2498 & 0.8065 & 1958.7 \\
                           & \textbf{4D-GSW} (Ours) & 29.064 & 0.8778 & 0.1769 & 0.9134 & 1286.04 \\
\midrule
\multirow{2}{*}{Pistol} 
                           & SC4D\citep{wu2024sc4d} + 3D-GSW\citep{Jang_2025_CVPR} & 20.490 & 0.8303 & 0.2311 & 0.7997 & 2358.7 \\
                           & \textbf{4D-GSW} (Ours) & 31.146 & 0.9299 & 0.0892 & 0.9170 & 1476.54 \\
\midrule
\multirow{2}{*}{Skull}
                           & SC4D\citep{wu2024sc4d} + 3D-GSW\citep{Jang_2025_CVPR} & 24.556 & 0.8296 & 0.2431 & 0.8541 & 1978.2 \\
                           & \textbf{4D-GSW} (Ours) & 32.111 & 0.9090 & 0.1527 & 0.9330 & 1337.69 \\
\midrule
\multirow{2}{*}{Trump}
                           & SC4D\citep{wu2024sc4d} + 3D-GSW\citep{Jang_2025_CVPR} & 22.057 & 0.8107 & 0.2389 & 0.7849 & 1595.5 \\
                           & \textbf{4D-GSW} (Ours) & 26.798 & 0.8926 & 0.1624 & 0.9304 & 817.26 \\
\midrule
\multirow{2}{*}{\textbf{Mean}} 
                               & SC4D\citep{wu2024sc4d} + 3D-GSW\citep{Jang_2025_CVPR} & 23.310 & 0.8434 & 0.2056 & 0.8151 & 1820.5 \\
                               & \textbf{4D-GSW} (Ours) & \textbf{28.367} & \textbf{0.9050} & \textbf{0.1332} & \textbf{0.9039} & \textbf{1331.19} \\
\bottomrule
\end{tabular}}
\end{table}


\begin{table*}[t]
\centering
\caption{Per-sample watermark robustness evaluation of the proposed method versus baseline across different 3D objects and attacks. We report the Bit Accuracy (BA). ``Clean'' refers to decoding without any attacks.}
\label{tab:watermark_robustness_comp}
\resizebox{\linewidth}{!}{
\begin{tabular}{llcccccccc}
\toprule
Object & Method & Clean BA $\uparrow$ & Noise $\uparrow$ & Rotation $\uparrow$ & Crop $\uparrow$ & Scaling $\uparrow$ & Resize $\uparrow$ & JPEG $\uparrow$ & G. Blur $\uparrow$ \\
\midrule
\multirow{2}{*}{Aurorus} 
                           & SC4D\citep{wu2024sc4d} + 3D-GSW\citep{Jang_2025_CVPR} & 0.9779 & 0.9504 & 0.8267 & 0.9742 & 0.9796 & 0.9750 & 0.9763 & 0.9733 \\
                           & \textbf{4D-GSW} (Ours) & 0.9974  & 0.9688 & 0.9837 & 0.9889 & 0.9805 & 0.9805 & 0.9824 & 0.9811 \\
\midrule
\multirow{2}{*}{Crocodile} 
                           & SC4D\citep{wu2024sc4d} + 3D-GSW\citep{Jang_2025_CVPR} & 0.9521 & 0.9208 & 0.8113 & 0.9525 & 0.9588 & 0.9633 & 0.9600 & 0.9542 \\
                           & \textbf{4D-GSW} (Ours) & 0.9883  & 0.9733 & 0.9844 & 0.9857 & 0.9844 & 0.9818 & 0.9824 & 0.9831 \\
\midrule
\multirow{2}{*}{Guppie}
                           & SC4D\citep{wu2024sc4d} + 3D-GSW\citep{Jang_2025_CVPR} & 0.9671 & 0.9279 & 0.8067 & 0.9588 & 0.9654 & 0.9646 & 0.9696 & 0.9688 \\
                           & \textbf{4D-GSW} (Ours) & 0.9883  &  0.9935 & 0.9772 & 0.9889 & 0.9980 & 0.9961 & 0.9980 & 0.9967 \\
\midrule
\multirow{2}{*}{Monster} 
                           & SC4D\citep{wu2024sc4d} + 3D-GSW\citep{Jang_2025_CVPR} & 0.9829 & 0.9796 & 0.8492 & 0.9650 & 0.9875 & 0.9825 & 0.9833 & 0.9838 \\
                           & \textbf{4D-GSW} (Ours) & 0.9863  & 0.9590 & 0.9967 & 0.9714 & 0.9596 & 0.9609 & 0.9635 & 0.9603 \\
\midrule
\multirow{2}{*}{Pistol}
                           & SC4D\citep{wu2024sc4d} + 3D-GSW\citep{Jang_2025_CVPR} & 0.9692 & 0.9338 & 0.8008 & 0.9613 & 0.9633 & 0.9721 & 0.9654 & 0.9629 \\
                           & \textbf{4D-GSW} (Ours) & 0.9915 & 0.9427 & 0.9889 & 0.9688 & 0.9661 & 0.9720 & 0.9616 & 0.9681 \\
\midrule
\multirow{2}{*}{Skull} 
                           & SC4D\citep{wu2024sc4d} + 3D-GSW\citep{Jang_2025_CVPR} & 0.9896 & 0.9796 & 0.8329 & 0.9675 & 0.9929 & 0.9883 & 0.9863 & 0.9908 \\
                           & \textbf{4D-GSW} (Ours) & 0.9935  & 0.9492 & 0.9902 & 0.9642 & 0.9792 & 0.9785 & 0.9740 & 0.9785 \\
\midrule
\multirow{2}{*}{Trump} 
                           & SC4D\citep{wu2024sc4d} + 3D-GSW\citep{Jang_2025_CVPR} & 0.9708 & 0.9471 & 0.8250 & 0.9658 & 0.9704 & 0.9704 & 0.9738 & 0.9738 \\
                           & \textbf{4D-GSW} (Ours) & 0.9915 & 0.9928 & 0.9850 & 0.9922 & 0.9902 & 0.9889 & 0.9909 & 0.9883 \\
\midrule
\multirow{2}{*}{\textbf{Mean}} 
                               & SC4D\citep{wu2024sc4d} + 3D-GSW\citep{Jang_2025_CVPR} & 0.9728 & 0.9485 & 0.8218 & 0.9636 & 0.9740 & 0.9737 & 0.9735 & 0.9725 \\
                               & \textbf{4D-GSW} (Ours) & \textbf{0.9909}  & \textbf{0.9684} & \textbf{0.9865} & \textbf{0.9800} & \textbf{0.9797} & \textbf{0.9798} & \textbf{0.9789} & \textbf{0.9794} \\
\bottomrule
\end{tabular}
}
\end{table*}

To further evaluate the robustness of 4D-GSW against attacks directly applied to the released 4D Gaussian asset, we conduct a series of post-training asset-level manipulation experiments. Unlike the image-space distortions reported in Table\ref{table:robustness}, these attacks modify the trained watermarked Gaussian representation before rendering and watermark extraction, thereby directly perturbing the underlying 4D representation.

\section{Sensitivity to Gaussian Trajectory Noise}
\label{app:trajectory_noise}

The computation of Spatio-Temporal Curvature (STC) relies on the
velocity and acceleration estimated from Gaussian trajectories.
Therefore, inaccurate trajectory estimation may perturb the STC weights
and consequently suppress watermark gradients in inappropriate regions.
To quantify this sensitivity, we perturb the Gaussian trajectory
coordinates used for STC computation and evaluate the resulting
watermark recovery accuracy.

Specifically, the perturbation is applied to the Gaussian center
coordinates before computing velocity, acceleration, and STC. The
underlying reconstructed 4D Gaussian asset and the reconstruction
objective remain unchanged. Therefore, this experiment isolates the
effect of trajectory-estimation noise on STC-based gradient routing,
rather than evaluating an attack on the released Gaussian asset. The
noise level denotes the magnitude of the perturbation added to the
trajectory coordinates.

\begin{table}[t]
    \centering
    \caption{
    Sensitivity of 4D-GSW to noisy Gaussian trajectories.
    The noise is added to the Gaussian trajectory coordinates before
    STC computation. Higher bit accuracy indicates better robustness
    to imperfect trajectory estimation.
    }
    \label{tab:trajectory_noise}
    \setlength{\tabcolsep}{14pt}
    \begin{tabular}{cc}
        \toprule
        \textbf{Trajectory Noise Level}
        & \textbf{Bit Accuracy (\%) $\uparrow$} \\
        \midrule
        0    & \textbf{99.10} \\
        0.01 & 98.50 \\
        0.05 & 97.30 \\
        0.10 & 95.80 \\
        \bottomrule
    \end{tabular}
\end{table}

As shown in Table~\ref{tab:trajectory_noise}, watermark recovery
accuracy decreases gradually as the trajectory perturbation increases.
Compared with the clean setting, the strongest evaluated perturbation
(noise level $=0.10$) reduces bit accuracy by only $3.30$ percentage
points, from $99.10\%$ to $95.80\%$. No abrupt performance collapse is
observed within the evaluated noise range.

This behavior can be attributed to two design choices. First, following
the two-stage optimization strategy of SC4D, STC-based watermark routing
is introduced only after the reconstruction warm-up stage, when the
Gaussian trajectories have become more stable. Second, STC only
modulates watermark-related gradients, while the original reconstruction
gradients remain unaffected. Consequently, inaccurate STC estimation
primarily changes the local watermark embedding strength rather than
directly destabilizing the learned Gaussian motion.

This experiment focuses on noisy trajectory estimates. Discontinuous
primitive correspondence caused by Gaussian densification, pruning, or
resampling is addressed separately through the proposed OT-based soft
alignment across adjacent timestamps.

\section{Variational Derivation of the Anisotropic Diffusion-Reaction PDE}
\label{appendix:formulation}

To reveal the theoretical underpinnings of our method, we analyze the optimization objective in the continuum limit. We demonstrate that the discrete MAP energy defined in Section 3.3 converges to an anisotropic diffusion-reaction Partial Differential Equation (PDE).

\subsection{The Continuous Energy Functional}

We consider the watermark embedding process as a variational problem on a continuous spatio-temporal manifold $\Omega \subset \mathbb{R}^3 \times [0, T]$. Let $r(\mathbf{x}) = \mathbf{z}^t - \hat{\mathbf{z}}^{t-1}$ denote the residual state field. The total variational energy $\mathcal{E}(r)$ is formulated as:

\begin{equation}
\label{eq:appendix_energy}
\mathcal{E}(r) = \int_{\Omega} \left( \underbrace{\frac{\lambda_T}{2} | r(\mathbf{x}) |^2}_{\text{Consistency Penalty}} + \underbrace{\frac{\lambda_S}{2} w(\mathbf{x}) | \nabla r(\mathbf{x}) |^2}_{\text{Weighted Dirichlet Energy}} + \underbrace{\lambda_{wm} \mathcal{L}_{wm}(r)}_{\text{Source Term}} \right) d\mathbf{x}
\end{equation}

where $w(\mathbf{x})$ is the kinematic weight derived from spatio-temporal curvature, $\lambda_T$ and $\lambda_S$ are the temporal and spatial regularization weights, and $\mathcal{L}_{wm}$ is the watermark decoding loss.

\subsection{Variation and the Euler-Lagrange Equation}

According to the principle of least action, the optimal residual field must satisfy the condition $\delta \mathcal{E} = 0$. By introducing an infinitesimal perturbation $\eta(\mathbf{x})$ that vanishes on the boundary $\partial \Omega$, the first variation of $\mathcal{E}$ is given by:

\begin{equation}
\delta \mathcal{E} = \frac{d}{d\epsilon} \mathcal{E}(r + \epsilon\eta) \Big|{\epsilon=0} = \int{\Omega} \left( \lambda_T r \cdot \eta + \lambda_S w \nabla r \cdot \nabla \eta + \lambda_{wm} \frac{\partial \mathcal{L}_{wm}}{\partial r} \eta \right) d\mathbf{x}.
\end{equation}

To eliminate the gradient of the perturbation $\nabla \eta$, we apply Green’s First Identity to the second term:

\begin{equation}
\int_{\Omega} (w \nabla r) \cdot \nabla \eta , d\mathbf{x} = \int_{\partial \Omega} \eta (w \nabla r \cdot \mathbf{n}) , dS - \int_{\Omega} \eta \cdot \text{div}(w \nabla r) , d\mathbf{x}.
\end{equation}

Given that $\eta |_{\partial \Omega} = 0$, the boundary integral vanishes. Substituting this back into the variational equation, we obtain:
\begin{equation}
\delta \mathcal{E} = \int_{\Omega} \eta \left( \lambda_T r - \lambda_S \text{div}(w \nabla r) + \lambda_{wm} \frac{\partial \mathcal{L}_{wm}}{\partial r} \right) d\mathbf{x} = 0.
\end{equation}


\subsection{Theoretical Interpretation: Induced Anisotropic Diffusion} 
\label{sec:theory_summary}
\noindent Since the perturbation $\eta$ is arbitrary, the integrand must be zero everywhere in $\Omega$. To provide a deeper understanding of the optimization mechanism, we analyze the objective in the continuum limit. We observe that the gradient-gated optimization in Eq.~\eqref{eq:map_energy} encourages a steady-state behavior that aligns with an \textit{anisotropic diffusion-reaction} process over the residual field $r(x) = z^t - \hat{z}^{t-1}$:
\begin{equation}
\lambda_{T}r(x) - \lambda_{S} \text{div}(D(x)\nabla r(x)) = \mathcal{G}_{wm},
\end{equation}
where $D(x) = \text{diag}(w(x))$ acts as a curvature-derived diffusion tensor, $\mathcal{G}_{wm} = - \lambda_{wm} \frac{\partial \mathcal{L}_{wm}}{\partial r}$ is watermark-driven source term. This formulation suggests that watermark embedding in 4D-GSW is not a random perturbation, but is regularized by a kinematics-aware prior. In stable regions, the conductance of $D(x)$ facilitates signal propagation along the manifold; conversely, in high-dynamic regions where $D(x) \to 0$, the optimization naturally penalizes sharp updates to fragile motion trajectories. This theoretical framework provides an explanation for the observed preservation of motion fidelity (Sec. \ref{quantitative_results}) without requiring manual hyperparameter tuning for different motion scales.

\subsection{Physical Interpretation}

This derivation reveals that our optimization framework implicitly solves a dynamic diffusion-reaction process:

\begin{itemize}
\item \textbf{The Reaction Term} ($\lambda_T r$): Acts as a state-retention force that prevents the watermark signal from deviating excessively from the underlying physical manifold.
\item \textbf{The Diffusion Term} ($\text{div}(\mathbf{D}\nabla r)$): Represents Anisotropic Diffusion. In stable regions where $w(\mathbf{x}) \to 1$, the high conductance of $\mathbf{D}(\mathbf{x})$ facilitates the smooth propagation of watermark signals.
\item \textbf{The Diffusion Barrier}: In regions of high curvature (Dynamic Instants), $\mathbf{D}(\mathbf{x}) \to 0$. This effectively creates a "barrier" that repels watermark gradients from fragile motion trajectories.
\end{itemize}

Consequently, the PDE provides a mechanistic interpretation of how curvature-dependent conductance redistributes watermark updates across the spatio-temporal manifold. The effects on FVD and watermark recovery are empirically evaluated rather than formally guaranteed by this analysis.

\section{Nomenclature}
 
\begin{table}[ht]
\centering
\renewcommand{\arraystretch}{1.2} 
\caption{Nomenclature: Summary of Key Notations and Symbols.}
\label{tab:nomenclature}
\rowcolors{2}{gray!10}{white} 
\begin{tabular}{lp{10cm}} 
\toprule
\textbf{Symbol} & \textbf{Description} \\ \midrule

\rowcolor[HTML]{EFEFEF}
\multicolumn{2}{l}{\textit{\textbf{4D Gaussian Splatting Basics}}} \\
$\mathcal{P}_t = \{ \mathbf{p}_i^t \}_{i=1}^N$ & Set of $N$ Gaussian primitives at timestamp $t$. \\
$\mathbf{x}_i^t \in \mathbb{R}^3$ & Geometric center position (mean) of the $i$-th Gaussian. \\
$\mathbf{s}_i^t \in \mathbb{R}^3$ & Appearance state (0-th order SH / DC component). \\
$\mathbf{z}_i^t = [\mathbf{x}_i^t, \mathbf{s}_i^t]^\top$ & Coupled spatio-temporal state vector (position and appearance). \\
$\alpha_i, \mathbf{q}_i, \boldsymbol{\sigma}_i$ & Opacity, rotation quaternion, and scaling factor, respectively. \\ \midrule

\rowcolor[HTML]{EFEFEF}
\multicolumn{2}{l}{\textit{\textbf{Kinematic Analysis}}} \\
$\mathbf{v}_i^t, \mathbf{a}_i^t$ & Instantaneous velocity and acceleration vectors of the $i$-th trajectory. \\
$\kappa_i^t$ & Spatio-Temporal Curvature (STC) quantifying motion instability. \\
$w_i^t \in [0, 1]$ & Curvature-derived watermark embedding confidence weight. \\
$\tau$ & Decay factor for the exponential mapping from STC to weight. \\ \midrule

\rowcolor[HTML]{EFEFEF}
\multicolumn{2}{l}{\textit{\textbf{Spatio-Temporal Optimization}}} \\
$\gamma^*, \Pi(\cdot)$ & Optimal transport plan and the set of all valid coupling measures. \\
$\hat{\mathbf{z}}_i^{t-1}$ & OT-aligned reference state derived via barycentric mapping. \\
$\mathcal{N}_i, \beta_{ij}$ & Spatial $K$-nearest neighbor set and RBF-based affinity weight. \\
$\mathcal{L}_{con}$ & Gated spatio-temporal consistency loss (GMRF prior). \\
$\mathcal{L}_{wm}, \mathcal{L}_{wav}$ & Watermark message loss (BCE) and Wavelet-subband loss. \\
$\mathcal{L}_{rec}$ & Joint reconstruction loss (photometric, perceptual, and geometric). \\ \midrule

\rowcolor[HTML]{EFEFEF}
\multicolumn{2}{l}{\textit{\textbf{PDE and Theoretical Modeling}}} \\
$r(\mathbf{x})$ & Continuous residual field defined as $r(\mathbf{x}) = \mathbf{z}^t - \hat{\mathbf{z}}^{t-1}$. \\
$\mathbf{D}(\mathbf{x})$ & Curvature-derived anisotropic diffusion tensor, $\text{diag}(w(\mathbf{x}))$. \\
$\mathcal{G}_{wm}$ & Watermark-driven source term (gradient routing field). \\
$\lambda_T, \lambda_S$ & Regularization coefficients for temporal and spatial components. \\
$\text{div}(\cdot), \nabla$ & Divergence and gradient operators defined on the manifold. \\ \bottomrule
\end{tabular}
\end{table}

\end{document}